\documentclass[]{fairmeta}

\DeclareUnicodeCharacter{2212}{-}
\title{Polarization-resolved imaging improves eye tracking}

\author[1]{Mantas \v{Z}urauskas}
\author[2]{Tom Bu}
\author[1]{Sanaz Alali}
\author[1]{Beyza Kalkanli}
\author[1,*]{Derek Shi}
\author[1]{Fernando Alamos}
\author[3]{Gauresh Pandit}
\author[1]{Christopher Mei}
\author[2]{Ali Behrooz}
\author[3]{Ramin Mirjalili}
\author[1]{Dave Stronks}
\author[1]{Alexander Fix}
\author[1]{Dmitri Model}

\affiliation[1]{Meta, Reality Labs, Redmond, WA 98052, USA}
\affiliation[2]{Meta, Reality Labs, Burlingame, CA 94010, USA}
\affiliation[3]{Meta, Reality Labs, Sunnyvale, CA 94089, USA}

\contribution[*]{Work done at Meta}

\abstract{Polarization-resolved near-infrared imaging adds a useful optical contrast mechanism to eye tracking by measuring the polarization state of light reflected by ocular tissues in addition to its intensity. In this paper we demonstrate how this contrast can be used to enable eye tracking.  Specifically, we demonstrate that a  polarization-enabled eye tracking (PET) system composed of a polarization--filter--array camera paired with a linearly polarized near-infrared illuminator can reveal trackable features across the sclera and gaze-informative patterns on the cornea, largely absent in intensity-only images. Across a cohort of 346 participants, convolutional neural network based machine learning models trained on data from PET reduced the median 95th-percentile absolute gaze error by 10--16\% relative to capacity-matched intensity baselines under nominal conditions and in the presence of eyelid occlusions, eye-relief changes, and pupil-size variation. These results link light--tissue polarization effects to practical gains in human--computer interaction and position PET as a simple, robust sensing modality for future wearable devices.}

\date{\today}
\correspondence{Mantas \v{Z}urauskas at \email{mantas@meta.com}}

\begin{document}

\maketitle

\section*{Introduction}

Eye tracking (ET) is foundational for interaction, rendering, and personalization in wearable systems \cite{plopski2022eye}. Conventional ET pipelines \cite{mansour2025enabling} built on intensity-only imaging---and on cues such as pupil localization and corneal glints---can degrade under eyelid/eyelash occlusions, eye-relief changes (device slippage), and variable pupil size \cite{hou2024unveiling}. To maintain accuracy, these systems often resort to multi-camera arrangements or complex in-field sensing architectures, increasing size, complexity, and cost---an undesirable trade-off for compact, always-on consumer devices.\cite{jin2024eye}

Polarization-sensitive imaging enables single-camera, single-illuminator architecture alternative that allows fewer components, lower power, and simplified calibration compared to multi-camera systems. The eye structure is primarily supported by a scaffold of birefringent fibrous collagen \cite{boote2020scleral,jarecki2024optimizing}. In practice, this transforms both cornea \cite{sobczak2023effect} and  sclera \cite{boote2020scleral} from regions that are largely featureless in grayscale into richly textured surfaces with dense, trackable features. Because these polarization-derived signals do not rely solely on pupil edges or engineered glints, they can sustain gaze inference when traditional cues are partially or wholly occluded. Polarization contrast adds a complementary sensing dimension by measuring the angle of linear polarization (AoLP) and degree of linear polarization (DoLP) of near-infrared (NIR) light reflected and scattered by ocular tissues. With a polarization–filter–array (PFA) camera \cite{rebhan2019principle} that integrates wire-grid micro-polarizers at the pixel level, per-pixel demosaicking produces AoLP/DoLP maps alongside total intensity.

This paper introduces polarization-enabled eye tracking (PET): a compact module that combines a polarization-filter-array (PFA) camera with a linearly polarized near-infrared (NIR) flood illuminator. Co-locating the camera and illuminator simplifies assembly and calibration while capturing polarization contrast from natural reflections and scattering in the eye. In practice, PET reveals fine scleral texture and repeatable corneal patterns that are difficult to see in intensity-only images. We observe these features consistently across a broad participant cohort and find that they remain stable over several weeks in a longitudinal study. On the algorithmic side, convolutional neural networks (CNN) trained directly on PET data achieve lower gaze-estimation error than otherwise identical models trained on intensity-only inputs, including under changes in wearing position (eye relief) or pupil size without re-calibration.

This work makes three contributions: (i) a polarization-sensitive ET concept with minimalist hardware comprised of  a single-camera and a single-illuminator per eye and a convolutional neural network based machine learning algorithm; (ii) experimental results that demonstrate lower gaze errors and improved consistency when training on PET data versus intensity alone under the same experimental constraints; and (iii) human-subject evidence of PET feature visibility and temporal stability. Taken together, these findings indicate that polarization contrast increases input information density for end-to-end gaze estimation, providing a new technological direction for ET in wearable devices.

\section*{Results}
\subsection*{Overview}

Eye tracking provides a fast, reliable signal of user intent and attention, enabling hands‑free controls on wearable devices. We anticipate that gaze inputs will form key pathway for user inputs in AI glasses and extended reality devices \cite{fernandes2025gaze} and reliability of this input will be important for ensuring good user experience. We find that in practice, user experience with eye-tracking is best correlated with the tail (or "worst case") performance over expected usage conditions, rather than an average error. Therefore, for each participant, we focus on 95th percentile, \(E_{95}\), of per-frame absolute gaze error across expected usage conditions, which captures tail behavior under transient degradations (e.g., eyelid occlusion, eye-relief shifts, pupil dilation). To aggregate population-wide, we report \(U_{50}E_{95}\): the median \(E_{95}\) error across all participants in the study. Compared to mean or median per-frame error, \(E_{95}\) emphasizes reliability at the tail, which more directly maps to interactive user experience where occasional large errors dominate perceived quality. 

We evaluated polarization-enabled eye tracking (PET) against intensity-only baseline using a single acquisition pipeline and common camera–illumination geometry. We also ensured that gaze inference CNN network capacity (number of parameters) and data collection protocol is matched. Data were collected on a non–form-factor (NFF) benchtop station with a polarization–filter–array (PFA) camera and a single linearly polarized 850\,nm illuminator operated in flood mode. The resulting dataset comprises \(n = 346\) participants, split into training (\(n = 198\)), and validation (up to \(n = 148\)).  Unless otherwise noted, results are reported on the subject-disjoint validation set.

From each raw polarization recording, we formed two input modalities: (i) PET inputs comprising four linear-polarization channels \((0^{\circ},\,45^{\circ},\,90^{\circ},\,135^{\circ})\), and (ii) a pseudo-intensity input obtained by averaging the four channels to emulate a polarization-insensitive camera. Two end-to-end models of identical architecture and training schedule (PETNet1) were trained per modality; for fairness, the pseudo-intensity image was duplicated across four channels to match PET’s input dimensionality and model capacity.
Our primary endpoint is \(E_{95}\), the 95th-percentile absolute gaze error, summarized population-wise as \(U_{50}E_{95}\): the median \(E_{95}\) error across all users in the validation set. Uncertainty in reported results was estimated with nonparametric bootstrapping (participant-level resampling); shaded envelopes in Fig.~\ref{fig:readout} indicate 90\% confidence intervals (CIs). We report results for two temporal camera placements (higher temporal and lower temporal) under three test conditions: (1) nominal donning with calibration, (2) eye‑relief changes (“slippage”) without re‑calibration, and (3) pupil‑size changes without re‑calibration. (Note: here temporal camera position refers to the placement of the eye-tracking camera near the temporal (outer) side of the eye, closer to the temple - see Fig.~\ref{fig:readout})

\subsection*{Feature stability, temporal consistency and universal visibility}
\begin{figure}[t]
    \centering
    \includegraphics[width=\linewidth]{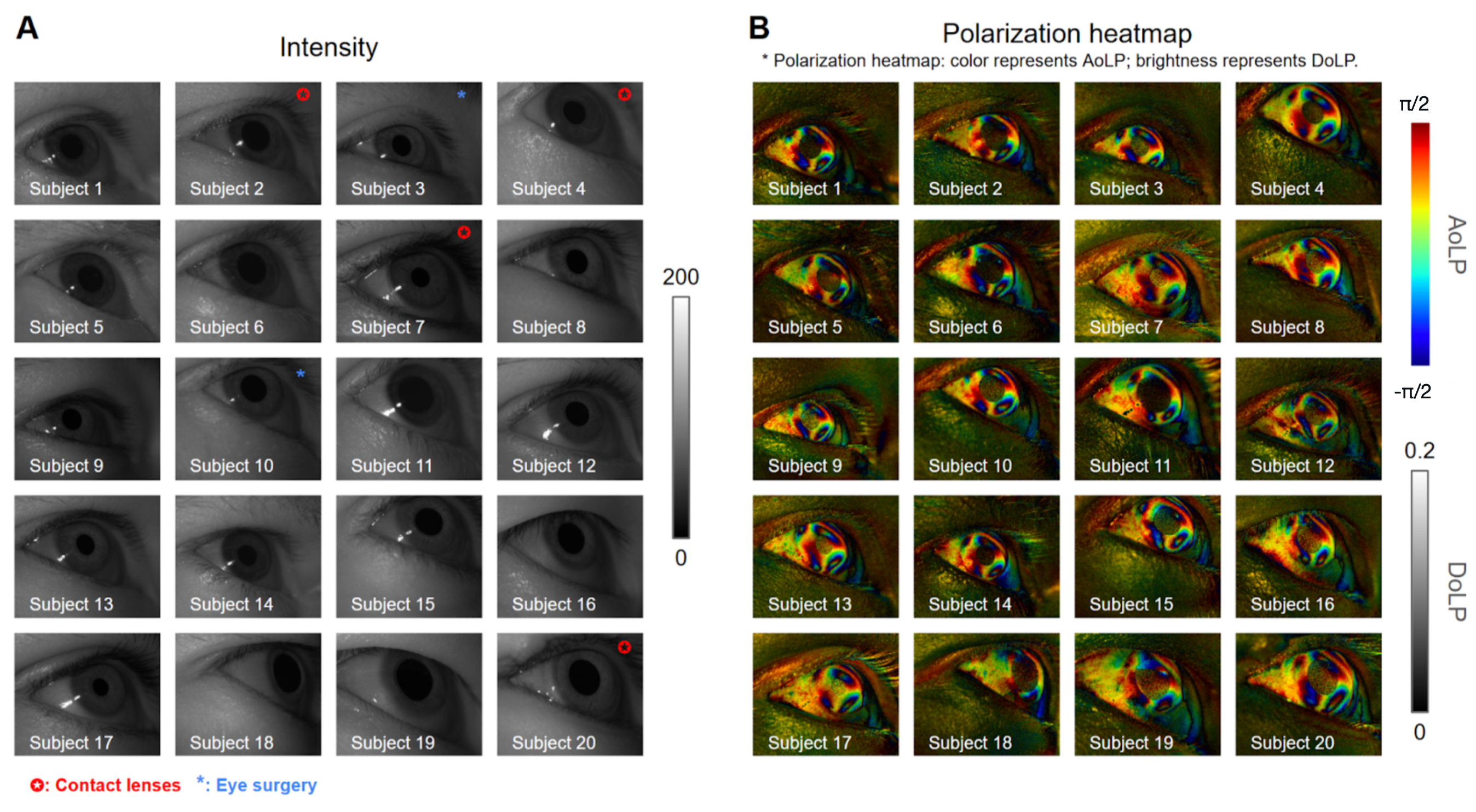}
    \caption{PET images with good cornea visibility across diverse subjects. The intensity images and the polarization heatmaps are shown in (A) and (B), respectively. Among the subjects, 4 subjects were wearing contact lenses during the data collection. And 2 participants had undergone laser eye surgeries (i.e., Subject 3 had SMILE, Subject 10 had LASIK)}
    \label{fig:Diversity}
\end{figure}

\begin{figure}[t]
    \centering
    \includegraphics[width=\linewidth]{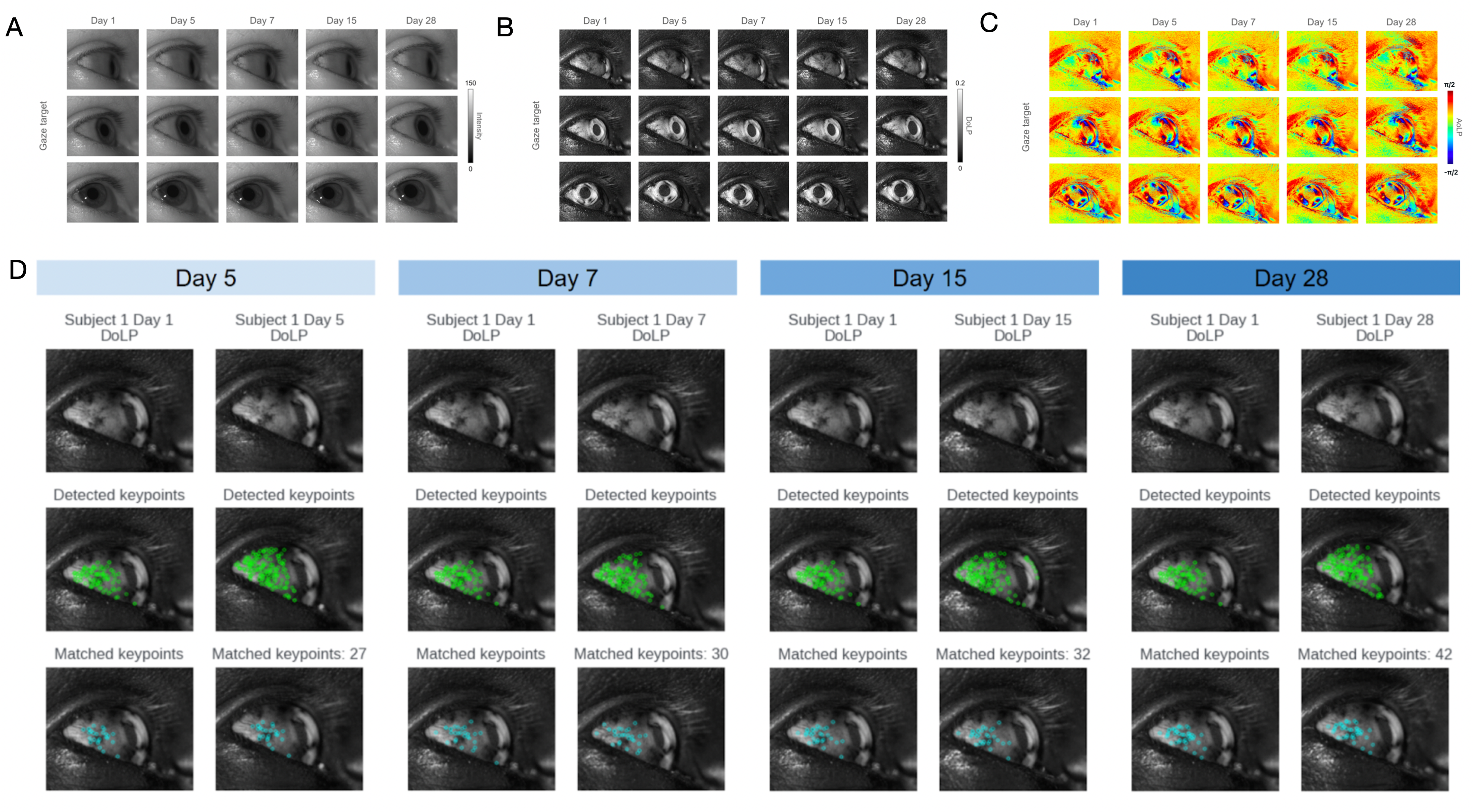}
    \caption{Polarization-resolved features over 4 weeks in one subject. (A) Intensity images. (B) DoLP. (C) AoLP. The persistence of polarization-derived scleral texture and corneal patterns across sessions support robust, personalized eye tracking and long‑term calibration.}
    \label{fig:stability}
\end{figure}

We assessed consistent presence of polarization-derived eye features in a 4‑week longitudinal study with a single volunteer imaged on Days 1, 5, 7, 15, and 28 using the same PET module and near‑coaxial camera–illuminator geometry (nominal donning). The sclera exhibited dense, fine‑grained polarization contrast that remained stable across all days based on qualitative and quantitative assessment; corneal polarization patterns were similarly repeatable. Composite AoLP/DoLP renderings shown in Fig.~\ref{fig:Diversity} revealed characteristic, subject‑specific texture fields that persisted despite minor variations in donning and background .
Quantitatively, scleral regions were matched across days using SIFT \cite{wu2013comparative} keypoints with geometric verification (RANSAC) \cite{derpanis2010overview}, consistently yielding inlier correspondences between sessions. With day 1 as a baseline matched keypoints on days 5, 7, 15 and 28 were respectively 27, 30, 32 and 42. This is consistent with expected noise ion re-positioning the participant in measuring rig and structural stability in healthy eyes. This stability suggests that brief user-specific calibration based on PET channels can remain valid over multi‑week horizons, reducing re‑calibration burden in consumer wearables and enabling persistent on‑device personal eye tracking calibration \cite{liu2024review}. The individuality and consistency of scleral patterns further indicate potential for integrated continuous authentication via the same sensing modality. Representative intensity, DoLP, and AoLP images across days are shown in Fig.~\ref{fig:stability}. 

Additionally we confirmed that similar features are visible across all participants. Fig.~\ref{fig:Diversity} demonstrates a subset of 20 volunteers including people with contact lenses and post-laser surgery.

\subsection*{Gaze tracking accuracy and robustness}
 Across nominal testing condition and both eye‑relief, and pupil‑size variations, PET lowers \(U_{50}E_{95}\) by \(0.12\!-\!0.23^{\circ}\) (10.3–15.9\%) relative to intensity‑only processing (Table~\ref{tab:e95u50_reduction}).:

\begin{table}[ht!]
\centering

\begin{tabular}{|p{2.1cm}|l|l|l|p{1.5cm}|p{1.5cm}|}
\hline
\textbf{Test condition} & \textbf{Camera position} & \textbf{PET \(U_{50}E_{95}\)} & \textbf{Intensity \(U_{50}E_{95}\)} & \textbf{Median reduction} & \textbf{Relative reduction} \\
 &  & \textbf{(deg; 90\% CI)} & \textbf{(deg; 90\% CI)} & \textbf{(deg)} & \textbf{(\%)} \\
\hline
Nominal (calibrated) & Lower temporal  & \(1.004^{\circ}\,[0.918\!-\!1.048]\) & \(1.126^{\circ}\,[1.017\!-\!1.153]\) & \(0.12^{\circ}\) & 10.8\% \\
\cline{2-6}
                     & Higher temporal & \(1.185^{\circ}\,[1.113\!-\!1.275]\) & \(1.395^{\circ}\,[1.289\!-\!1.474]\) & \(0.21^{\circ}\) & 15.0\% \\
\hline
Eye‑relief change (no recal.) & Lower temporal  & \(1.192^{\circ}\,[1.106\!-\!1.308]\) & \(1.359^{\circ}\,[1.264\!-\!1.449]\) & \(0.17^{\circ}\) & 12.2\% \\
\cline{2-6}
                              & Higher temporal & \(1.551^{\circ}\,[1.380\!-\!1.627]\) & \(1.747^{\circ}\,[1.638\!-\!1.879]\) & \(0.20^{\circ}\) & 11.2\% \\
\hline
Pupil‑size change (no recal.) & Lower temporal  & \(1.021^{\circ}\,[0.900\!-\!1.057]\) & \(1.138^{\circ}\,[1.019\!-\!1.160]\) & \(0.12^{\circ}\) & 10.3\% \\
\cline{2-6}
                              & Higher temporal & \(1.199^{\circ}\,[1.075\!-\!1.247]\) & \(1.426^{\circ}\,[1.272\!-\!1.448]\) & \(0.23^{\circ}\) & 15.9\% \\
\hline
\end{tabular}
\caption{\label{tab:e95u50_reduction} Population‑level tail gaze error with and without polarization contrast. Median 95th‑percentile absolute gaze error (\(U_{50}E_{95}\); degrees) for polarization‑enabled eye tracking (PET) versus intensity‑only processing under matched hardware and model capacity, reported across two temporal camera placements and three test conditions. Values are medians over participants with 90\% bootstrap confidence intervals (CIs). “Median reduction” = median(Intensity) − median(PET); “Relative reduction” = 100 × Median reduction ÷ median(Intensity). “No recal.” indicates no re‑calibration for the corresponding test condition.
}
\end{table}

\begin{figure}[t]
    \centering
    \includegraphics[width=\linewidth]{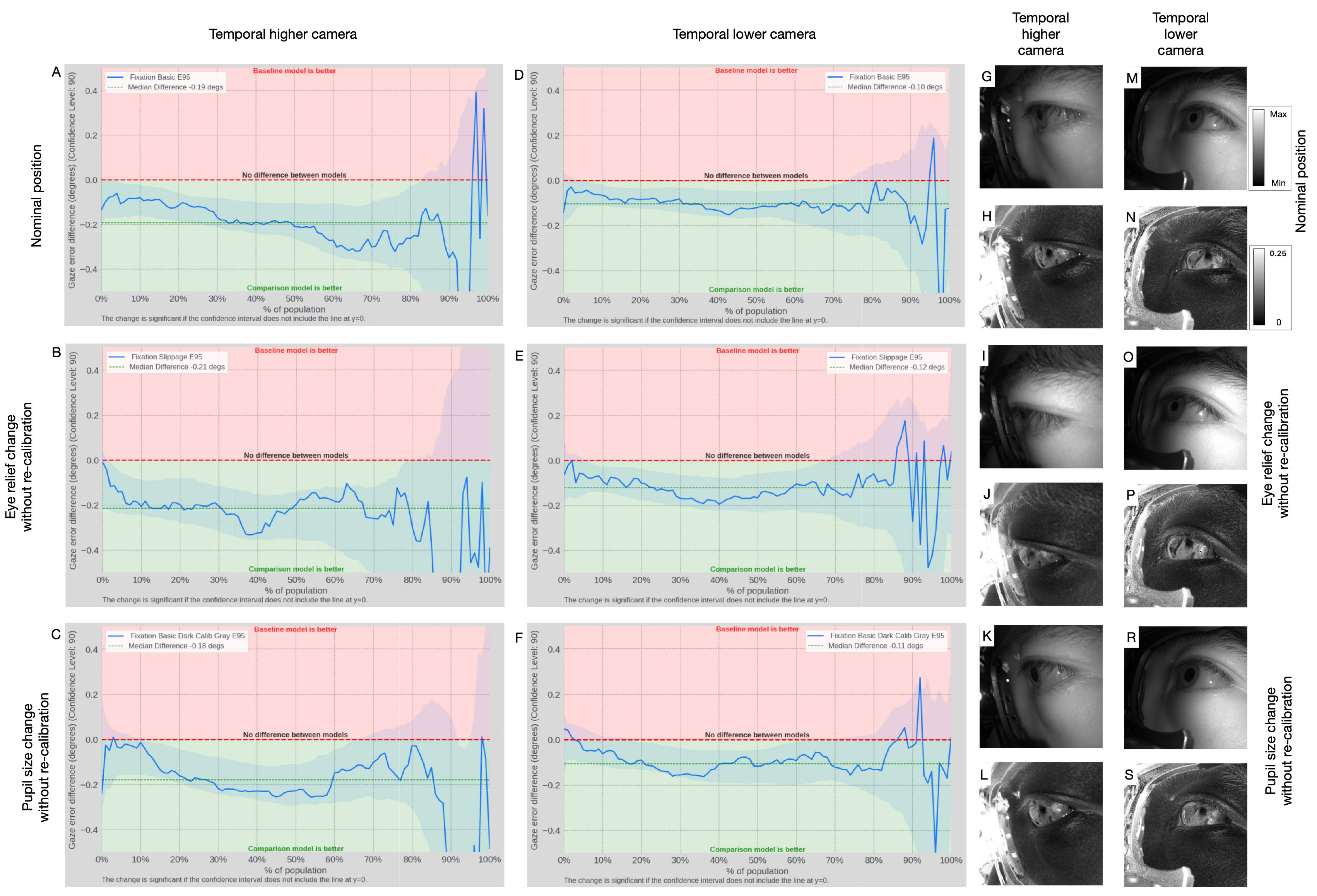}
    \caption{Intensity-only versus polarization-enabled eye tracking under matched acquisition and training. From identical raw polarization recordings, we formed (i) pseudo‑intensity inputs (superpixel averaging) and (ii) four‑channel polarization inputs \((0^{\circ}, 45^{\circ}, 90^{\circ}, 135^{\circ})\), trained capacity‑matched models, and compared population‑wise \(U_{50}E_{95}\) across percentiles. \textbf{A–F}, Median PET–intensity difference (blue) with 90\% bootstrapped confidence envelopes for higher/lower temporal placements under three test conditions: nominal, eye‑relief change (no re‑calibration), and pupil‑size change (no re‑calibration). Shaded bands annotate regimes of lower error for each modality. \textbf{G–S}, Representative input pairs (pseudo‑intensity vs polarization‑resolved) by condition.}
    \label{fig:readout}
\end{figure}

Population-difference curves (median traces with 90\% CIs) show that the bootstrapped CI for the PET–intensity median difference excludes zero across broad percentile ranges in all three conditions and both placements (Fig.~\ref{fig:readout}A–F), indicating statistically significant improvements. Qualitatively, PET sustains its advantage when traditional cues degrade (slippage, pupil dilation), consistent with the hypothesis that polarization contrast increases input information density for end‑to‑end learning in a single‑camera, single‑illuminator configuration.
\noindent\textbf{Baseline parity.} The intensity baseline uses identical hardware, acquisition geometry, and raw sensor data as PET. The only difference is input formation (four-channel polarization stack for PET versus channel-averaged pseudo-intensity duplicated to four channels). Models share the same backbone, head, and training schedule, ensuring a fair comparison that isolates the contribution of polarization contrast.
\noindent\textbf{Statistical assessment.} Participant-level bootstrapping (resampling users with replacement and recomputing \(U_{50}E_{95}\)) was used to derive 90\% CIs for median differences. Across conditions and placements, these CIs exclude zero over wide percentile spans (Fig.~\ref{fig:readout}A–F), supporting population-level gains with PET under matched acquisition and training.

\section*{Discussion}

We demonstrated that polarization-sensitive imaging increases the effective information available to eye-tracking systems and translates into consistent, population‑level accuracy gains under matched hardware and training capacity. Using a polarization–filter–array camera and a single linearly polarized NIR illuminator, polarization‑enabled eye tracking (PET) exposes dense scleral texture and repeatable corneal polarization patterns that are largely absent in intensity‑only inputs. Under identical acquisition, models trained on PET channels reduced population‑wise error (\(U_{50}E_{95}\)) by \(\sim 10\!-\!16\%\) across nominal and non-ideal conditions, and sustained their advantage when traditional cues degraded (eye‑relief changes, pupil‑size variation; Fig.~\ref{fig:readout}, Table~\ref{tab:e95u50_reduction}). These findings support the central hypothesis that polarization contrast boosts gaze‑informative content for end‑to‑end learning without resorting to multi‑camera geometry \cite{wang2025accurate} or structured illumination \cite{zhang2025fringe}.

We observed that PET-related performance gains are consistently larger for the higher temporal camera than for the lower temporal camera (Fig.~\ref{fig:readout}, Table~\ref{tab:e95u50_reduction}). The higher temporal placement affords reduced visibility of the eye opening and more frequent eyelid/eyelash occlusions, which disproportionately suppress conventional intensity cues (e.g., pupil contours and corneal glints). Practically, this suggests that PET may be compatible with constrained, industrially realistic module placements, where frames, brow, or optical packaging limit aperture.

The observed gains have a clear optical and physiological basis. The sclera’s anisotropic collagen under multiple scattering imprints a non‑zero degree and a stable angle of linear polarization, yielding fine‑scale, temporally consistent contrast. At the cornea, both interface optics (Fresnel reflection/refraction at the air–tear film–cornea stack) and birefringence of the organized stromal lamellae contribute: the former generates specular and refracted highlights, while the latter introduces phase retardance rotation of the polarization state. The resulting AoLP/DoLP patterns augment pupil contours and conventional glints. Together, these polarization‑resolved cues broaden the effective feature set, reduce single‑cue brittleness, and maintain accuracy with changing eye pupil sizes or modest eye‑relief shifts. 

It is important that these benefits can be harnessed with a single camera and a single linearly polarized NIR illuminator, enabling compact optical assemblies that are directly compatible with modularization for the purpose of wearable deployment. While we have accounted for design limitations trade‑offs in this study by penalizing PET, ideally they could be refined through future developments of camera modules. (i) PFA mosaics exchange angular sampling for per‑pixel resolution; demosaicking can reduce per‑channel SNR. (ii) Polarization efficiency and measured AoLP/DoLP depend on incidence angle and relative sensor–illuminator orientation; off‑axis geometry and uncontrolled reflections can compress contrast. (iii) Our results were obtained on a non–form‑factor benchtop station; porting gains to compact optics will require attention to stray polarization, coatings, stray light suppression, and mechanical tolerances. (iv) We used a single NIR wavelength and linear polarization; temporal multiplexing of linear states or circular states may be explored to improve robustness across ocular phenotypes and ambient conditions but may introduce new emitter/filter complexity and power trade‑offs.

These constraints outline several paths forward. On the sensing side, miniaturized polarimetric camera sensors with low interpixel cross-talk, high optical throughput and fill factor are required for module miniaturization. Alternatively polarization contrast could be provided using metasurface routers at the sensor \cite{soma2024metasurface, zuo2023chip} or module level \cite{rubin2019matrix}. On the algorithmic side, architectures that explicitly model personalization (e.g., lightweight user‑specific adapters or priors over eye geometry) \cite{liu2024review} are expected to compound PET’s gains while keeping calibration minimal, and self‑supervised objectives could leverage the structured relationships among AoLP, DoLP, and intensity.

Beyond gaze estimation, PET provides access to birefringence‑linked contrast: organized collagen lamellae in the cornea and anisotropic collagen bundles in the sclera rotate and differentially retard NIR polarization, revealing meso‑scale collagen “scaffolding” that is largely invisible to intensity imaging \cite{boote2020scleral}. Clinical studies indicate that structural changes may correlate with various pathologies such as myopia progression \cite{liu2023posterior}, keratoconus \cite{fukuda2013keratoconus, bui2023keratoconus}, post LASIK ectasia \cite{bueno2006corneal, bohac2018incidence}.  If validated in larger, longitudinal cohorts, trend‑based analyses of this scaffolding could support new health applications on consumer devices.

In summary, polarization contrast offers a simple, physically grounded route to enrich the visual cues available for end‑to‑end gaze estimation. The resulting improvements in accuracy, robustness to donning and pupil‑size variation, and multi‑week stability—achieved with a single camera and single illuminator—expand the operating envelope of eye tracking in compact wearables. By tying algorithmic gains to photonic signal formation and outlining clear engineering levers for integration, PET provides a practical path toward reliable, low‑burden eye‑based input for human–computer interaction.

\section*{Methods}
\subsubsection*{Polarization sensitive imaging}

\begin{figure}[t]
    \centering
    \includegraphics[width=\linewidth]{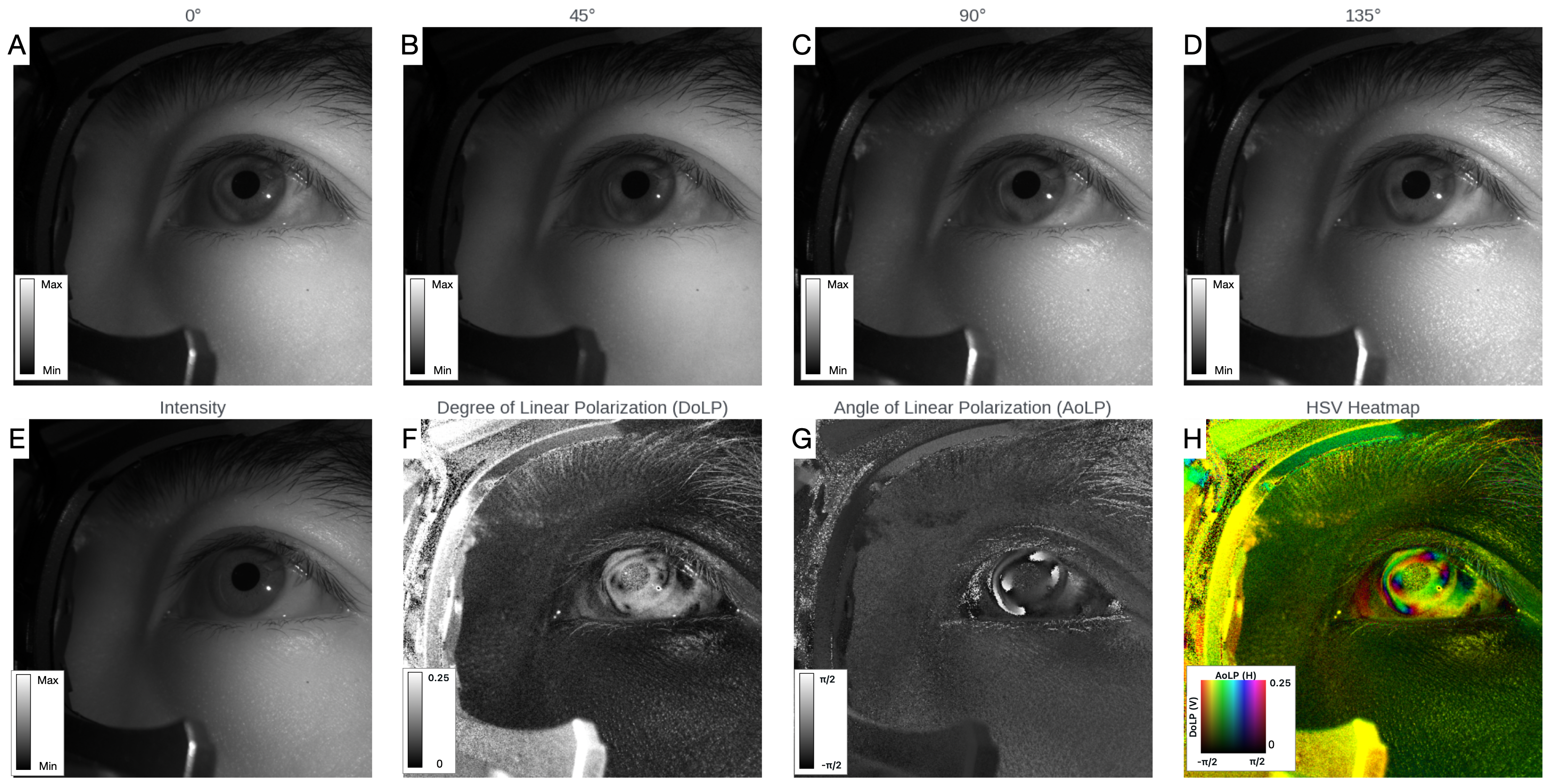}
    \caption[Polarization-resolved imaging of the human eye]{Polarization-resolved imaging of the human eye. Panels show (a--d) the four reconstructed linear polarization channels at \(0^\circ\), \(45^\circ\), \(90^\circ\), and \(135^\circ\) (grayscale); (e) total intensity \(I = S_0/4\); (f) degree of linear polarization \(\mathrm{DoLP} = \sqrt{S_1^2 + S_2^2}/(S_0 + \varepsilon)\) clipped to \([0, 1]\); (g) angle of linear polarization \(\mathrm{AoLP} = \tfrac{1}{2}\,\arctan2(S_2, S_1)\) (radians); and (h) an HSV composite where hue encodes AoLP, saturation scales with DoLP, and value is the gamma-corrected intensity. All panels are rotated by \(180^\circ\) and contrast-normalized for visualization.}
    \label{fig:pol_eye}
\end{figure}

A polarization-sensitive imaging system with a micro-polarizer mosaic records raw intensities at four linear orientations (\(0^\circ, 45^\circ, 90^\circ, 135^\circ\)). The raw image is demosaicked to reconstruct full-resolution images for each polarization orientation, then smoothed with a Gaussian filter (\(\sigma = 1\)). The Stokes parameters are computed as \(S_0 = I_{0^\circ} + I_{45^\circ} + I_{90^\circ} + I_{135^\circ}\), \(S_1 = I_{0^\circ} - I_{90^\circ}\), and \(S_2 = I_{45^\circ} - I_{135^\circ}\). Total intensity is \(I = S_0/4\); the degree of linear polarization is \(\mathrm{DoLP} = \sqrt{S_1^2 + S_2^2}/(S_0 + \varepsilon)\) with a small \(\varepsilon\) for numerical stability; and the angle of linear polarization is \(\mathrm{AoLP} = \tfrac{1}{2}\,\arctan2(S_2, S_1)\) (radians). Pixels with very low \(S_0\) are masked in the DoLP and AoLP maps to suppress artifacts. For visualization, \(I\) is normalized by its 99th percentile, and an HSV composite maps AoLP to hue, and DoLP to value, with saturation = 1. For higher temporal camera panels, they are rotated by \(180^\circ\) to match the acquisition orientation and contrast-normalized for display. Examples of individual channels and derivative computed images are shown in Fig.~\ref{fig:pol_eye}

% Figure environment (you can replace \includegraphics with your figure)

\subsection*{Data Collection Set-up}
\begin{figure}[ht]
  \centering
  \includegraphics[width=\linewidth]{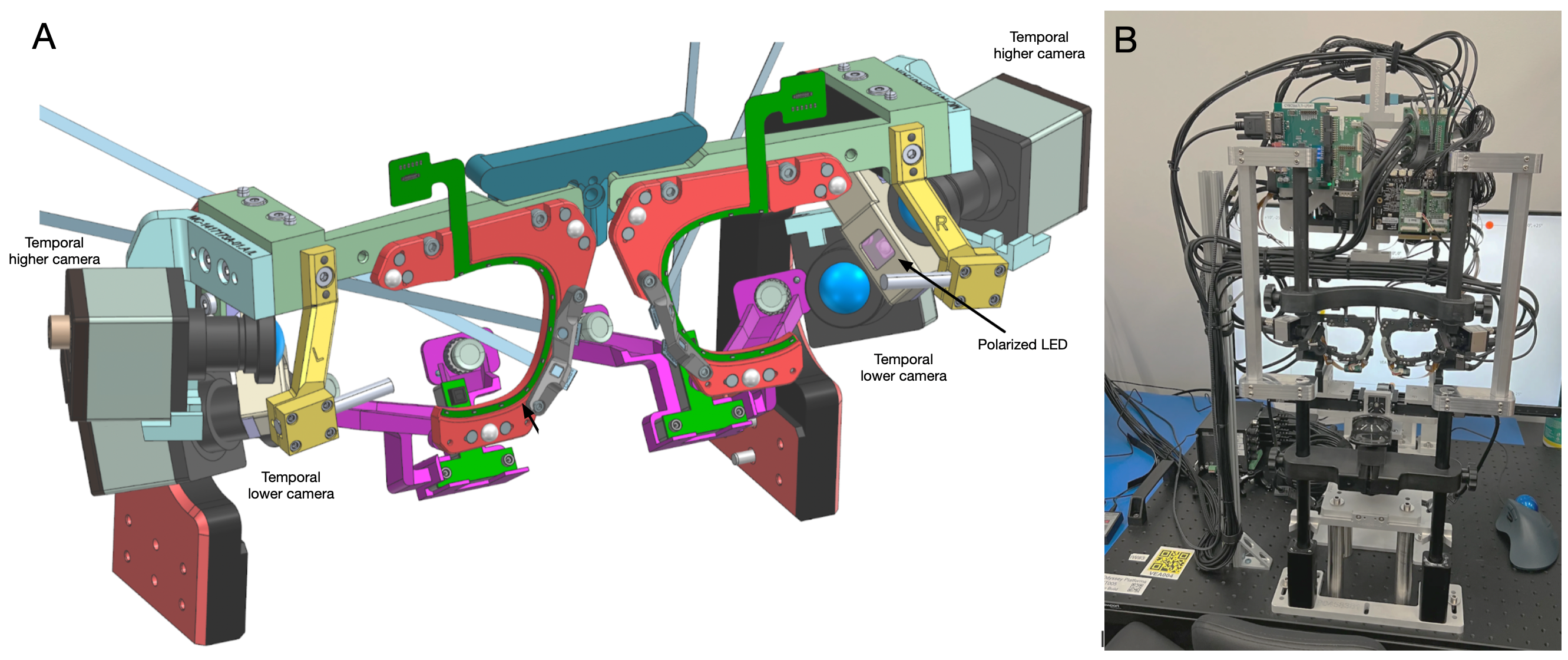}
  \caption{Data collection hardware. (\textbf{A}) PET subsystem and polarized 850\,nm flood illuminator integrated in an NFF benchtop station; polarization-sensitive cameras and a single-camera/illuminator PET concept. (\textbf{B}) Photograph of the custom in-house station with chinrest and adjustable kinematics for ER/IPD coverage.}
  \label{fig:system}
\end{figure}

Experiments were conducted on a benchtop station data collection platform shown in Fig.~\ref{fig:system}. The PET subsystem pairs a single polarization-filter-array (PFA) camera (IDS Imaging UI-3080CP-M-GL Rev.2 camera with Sony IMX250) with a near-infrared (NIR) illuminator (Osram LZ1-00R402) coupled with a wire-grid polarizer film (TECHSPEC-Edmund optics). In this benchtop configuration, the camera and illuminator were spatially separated. The setup was binocular, and for each eye, two cameras were mounted on the temporal side to capture distinct perspectives: higher and lower temporal. This arrangement enabled evaluation across perspective baselines while remaining faithful to the single-camera PET concept during modeling and analysis.
The PFA sensor was a Sony IMX250 (wire-grid), which uses a micro-polarizer mosaic to record raw intensities at four linear orientations \((0^\circ, 45^\circ, 90^\circ, 135^\circ)\). Per-pixel reconstruction yields total intensity (Stokes \(S_0\)) together with the Angle of Linear Polarization (AoLP) and the Degree of Linear Polarization (DoLP) channels.
For illumination, we used a single polarized 850\,nm LED with an integrated linear polarizer operated in \emph{flood mode}. Flood illumination avoids structured glints and leverages polarization contrast in light scattered from ocular tissues, enhancing features on the sclera and cornea.
\subsubsection*{Data Collection Station and Kinematics}
The camera–illumination pair was integrated into a custom, in-house benchtop station designed to maintain consistent image quality across eye relief (ER) and viewing geometry across participants while allowing controlled variation in donning and illumination background conditions. Participants were stabilized using a chinrest, and their eye positions were adjusted to sample a wide distribution of ER  for robust training coverage. We targeted a broad ER distribution and included short–eye-relief (slippage) sequences down to the minimum physically achievable ER (\(\geq 6\) mm). IPD was measured per participant and recorded as metadata.
\subsubsection*{Gaze Stimuli and Display Geometry}
Gaze targets were presented on a monitor positioned at 48 cm from the subject’s eye. The display proxy (monitor center) was aligned such that it corresponded to a \(\mathrm{-9.7^\circ}\) tilt relative to nominal \(0^\circ\) gaze. The data-collection field of view (FOV) covered \(\mathrm{30^\circ \times 20^\circ}\) (horizontal \(\times\) vertical) relative to \(0^\circ\) gaze. To mitigate occasional occlusions, the research assistant was permitted to slightly adjust target positions across all sequences. Sequences were captured under multiple conditions, including nominal donning with white or dark backgrounds and non-nominal donning with measured slippage to probe sensitivity to ER changes.

\subsection*{Gaze sequences}
All participants completed a battery of visual target sequences used in both training and evaluation splits. Unless otherwise noted, target patterns were centered at the monitor origin point \((0,0)\), corresponding to a \(-9.7^\circ\) tilt relative to nominal \(0^\circ\) gaze. To mitigate occasional target occlusions, the research assistant (RA) was permitted to slightly adjust target positions during data collection; such adjustments were allowed across all sequences presented in table shown in Table~\ref{tab:sequences}.

Under nominal donning with a neutral face and white background, participants performed:
(i) \texttt{NW\_RING20} (Calib; 9 points at \(20^\circ\), circle order; fixation 1\,s, transition 1\,s),
(ii) \texttt{NW\_RS} (Critical; Eval—Saccade and ML Training; 20 points randomly distributed within \(30^\circ \times 20^\circ\) FOV, random order; fixation 1\,s, transition 1\,s), and
(iii) \texttt{NW\_FP18} (Critical; Eval—Gaze Angle and ML Training; 18 points comprising an oval at \(30^\circ \times 20^\circ\) plus a \(10^\circ\) circle, oval order; fixation 1\,s, transition 1\,s).

For ML training under nominal donning with varied facial expressions on a white background, participants completed \texttt{NW\_RS\_SQUINT}, \texttt{NW\_RS\_BLINK}, and \texttt{NW\_RS\_EXPR} (each: 20 random points within \(30^\circ \times 20^\circ\), random order; fixation 1\,s, transition 0.1\,s).

\begin{figure}[ht]
  \centering
  \includegraphics[width=\linewidth]{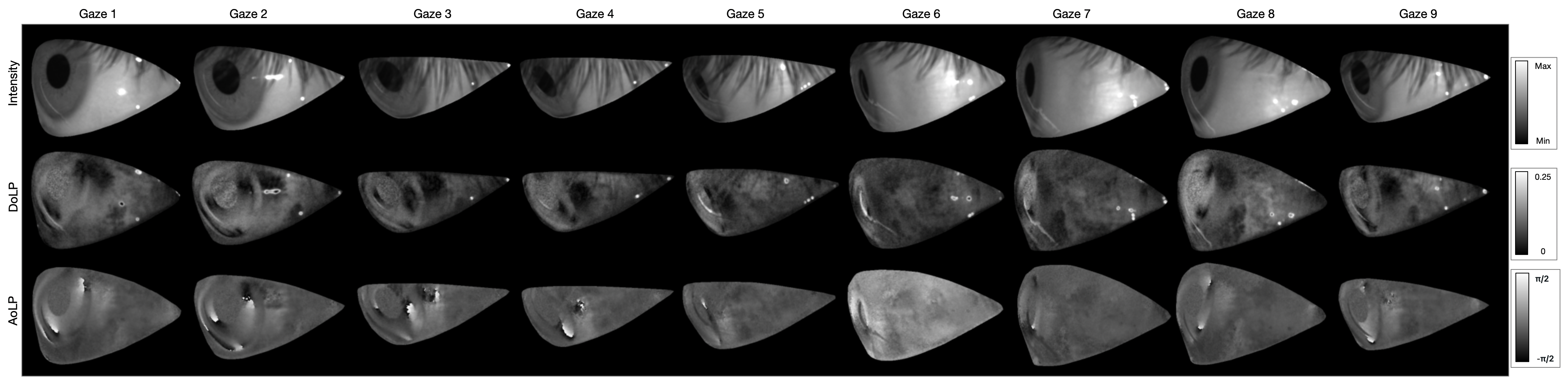}
  \caption{Sample images of segmented out eye opening at nine different gaze targets from nominal calibration sequence.}
  \label{fig:nominal}
\end{figure}

Under nominal donning with a dark background, participants performed \texttt{ND\_RING20} (Calib; 9 points at \(20^\circ\), circle order; fixation 1\,s, transition 1\,s) and \texttt{ND\_RS} (Eval; 20 random points within \(30^\circ \times 20^\circ\); fixation 1\,s, transition 0.1\,s). Example of 9 images of an eye in nominal donning position is shown in Fig.~\ref{fig:nominal}. 

To assess variations in donning, we included non‑nominal donning (measured slippage) with a neutral face and white background: \texttt{SW\_RS} (Critical; Eval and ML Training; 20 random points within \(30^\circ \times 20^\circ\), random order; fixation 1\,s, transition 0.1\,s), \texttt{SW\_RING20} (Critical; Calib; 9 points at \(20^\circ\), circle order; fixation 1\,s, transition 1\,s), and \texttt{SW\_FP18} (Eval and ML Training; 18 points: oval \(30^\circ \times 20^\circ\) + circle \(10^\circ\); oval order; fixation 1\,s, transition 1\,s).

Finally, under nominal donning with a neutral face and varying brightness backgrounds, participants completed \texttt{NBV\_RS} (Eval and ML Training; 20 random points within \(30^\circ \times 20^\circ\), random order; fixation 1\,s, transition 0.1\,s).

\begin{table}[ht]
\centering
\begin{tabular}{|p{3.2cm}|p{4.0cm}|p{1.8cm}|p{5.2cm}|}
\hline
\textbf{Condition} & \textbf{Sequence Name} & \textbf{\# of targets} & \textbf{Arrangement; order} \\
\hline
Nominal; Neutral; White BG & \texttt{NW\_RING20} (Critical) & 9 & Circular @ \(20^\circ\); circle \\
\hline
 & \texttt{NW\_RS} (Critical) & 20 & Random in FOV (\(30^\circ \times 20^\circ\)); random \\
\hline
 & \texttt{NW\_FP18} (Critical) & 18 & Oval \(30^\circ \times 20^\circ\) + Circle \(10^\circ\); oval \\
\hline
Nominal; Var. expressions; White BG & \texttt{NW\_RS\_SQUINT} & 20 & Random in FOV (\(30^\circ \times 20^\circ\)); random \\
\hline
 & \texttt{NW\_RS\_BLINK} & 20 & Random in FOV (\(30^\circ \times 20^\circ\)); random \\
\hline
 & \texttt{NW\_RS\_EXPR} & 20 & Random in FOV (\(30^\circ \times 20^\circ\)); random \\
\hline
Nominal; Dark BG & \texttt{ND\_RING20} & 9 & Circular @ \(20^\circ\); circle \\
\hline
 & \texttt{ND\_RS} & 20 & Random in FOV (\(30^\circ \times 20^\circ\)); random \\
\hline
Non‑nominal (Measured slippage); Neutral; White BG & \texttt{SW\_RS} (Critical) & 20 & Random in FOV (\(30^\circ \times 20^\circ\)); random \\
\hline
 & \texttt{SW\_RING20} (Critical) & 9 & Circular @ \(20^\circ\); circle \\
\hline
 & \texttt{SW\_FP18} & 18 & Oval \(30^\circ \times 20^\circ\) + Circle \(10^\circ\); oval \\
\hline
Nominal; Neutral; Varying BG & \texttt{NBV\_RS} & 20 & Random in FOV (\(30^\circ \times 20^\circ\)); random \\
\hline
\end{tabular}
\caption{\label{tab:sequences}
Summary of testing sequences (all used for both training and evaluation). Targets are centered at monitor \((0,0)\) (\(-9.7^\circ\) tilt from nominal \(0^\circ\) gaze).  Naming scheme: \emph{Context prefix}—NW (Nominal, neutral face, white BG), ND (Nominal, neutral face, dark BG), SW (Measured slippage, neutral face, white BG), NBV (Nominal, neutral face, varying brightness BG); \emph{Base sequence}—RING20 (9‑point ring at \(20^\circ\)), RS (Random Saccade in \(30^\circ \times 20^\circ\) FOV), FP18 (18‑point fixed pattern: oval \(30^\circ \times 20^\circ\) + circle \(10^\circ\)); \emph{Modifiers}—SQUINT, BLINK, EXPR (facial‑expression variants).
}
\end{table}

\subsection*{Machine learning model, PETNet1}

\begin{figure}[ht]
  \centering
  \includegraphics[width= 12 cm]{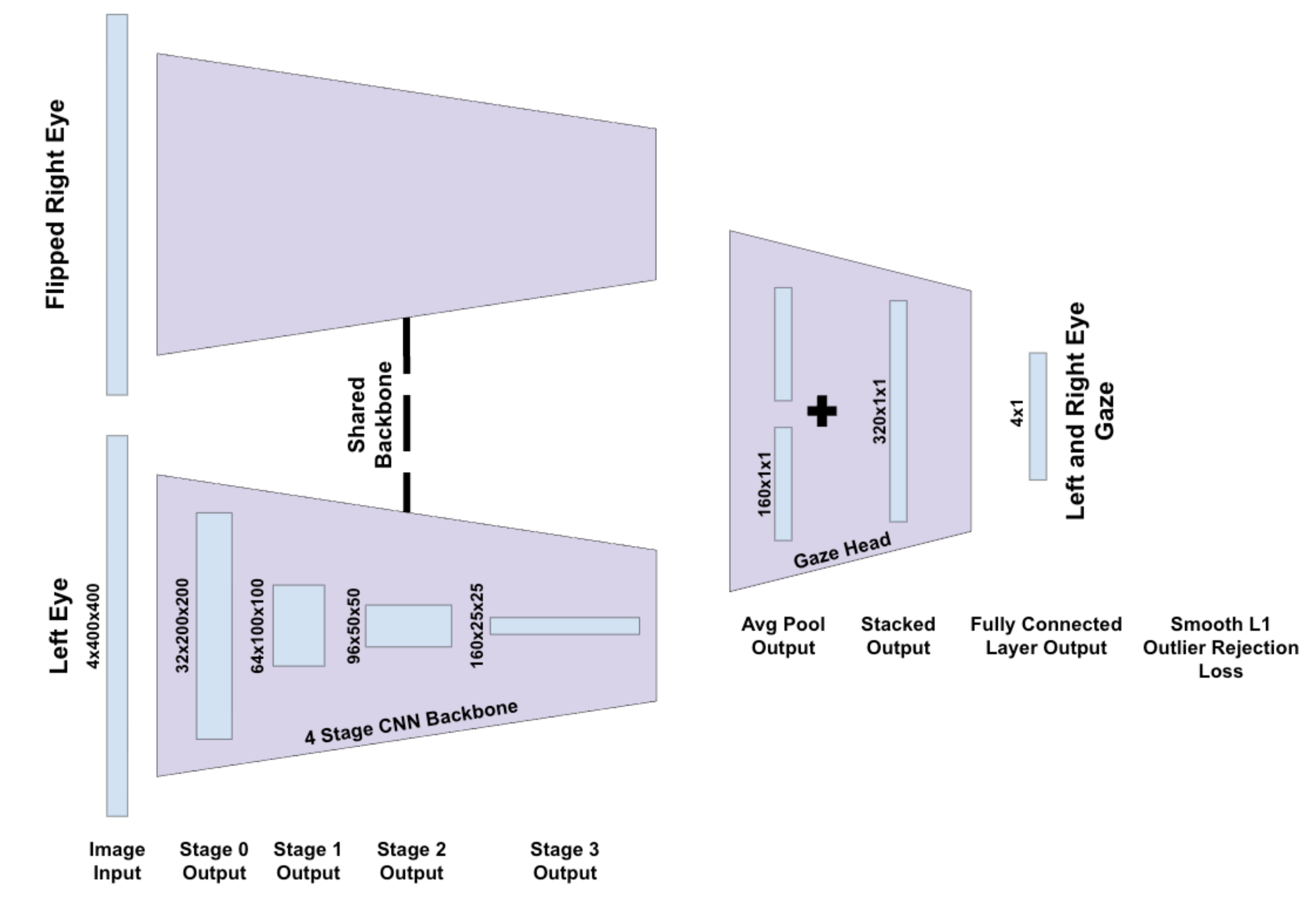}
  \caption{ Structural diagram of multi-view gaze regression model, which accepts images from both eyes. The images are passed through a 4-stage CNN backbone and then fed to a gaze regression head which predicts the gaze for each eye.}
  \label{fig:algo}
\end{figure}

We train an end-to-end convolutional model, PETNet1 to regress binocular gaze direction from polarization-sensitive eye images. The model architecture, presented in in Fig.~\ref{fig:algo} is extending previously reported CNN network \cite{mansour2025enabling}. It predicts per-eye gaze as polar angles relative to a fixed eye origin and a calibrated display geometry. A lightweight user calibration provides an affine correction (per-eye scale and bias) learned from a 9‑point sequence and applied at inference; unless otherwise stated, calibration is performed under nominal donning and held fixed for tests involving eye‑relief (“slippage”) and pupil‑size changes.

\textbf{Inputs.} Each raw frame yields four linear‑polarization images at \(0^\circ, 45^\circ, 90^\circ, 135^\circ\) via the micro‑polarizer mosaic. For the polarization‑enabled modality (PET), these four images are stacked channel‑wise per eye. For the intensity‑only baseline, the four orientations are averaged to form a pseudo‑intensity image, which is duplicated across four channels to match PET’s input dimensionality and model capacity. Per‑eye inputs are per‑channel normalized. Unless otherwise noted, synchronized binocular inputs are used and models are trained/evaluated separately for each camera placement (higher temporal, lower temporal).

\textbf{Architecture and fusion.} Each eye is processed by the same four‑stage backbone (shared weights), composed of inverted‑residual depthwise‑separable 3\(\times\)3 blocks. With a total stride of 16, the backbone produces \(25\times 25\) feature maps with 160 channels per eye; binocular features are concatenated to 320 channels before the prediction head. The backbone has \(\sim\)1.5M parameters. Convolutions are bias‑free with batch normalization; ReLU activations are used throughout. No cross‑view attention or specialized multi‑view modules are employed.

\textbf{Prediction head and personalization.} A shallow head maps the fused features to per‑eye gaze angles. Personalization is limited to the affine correction (two scales and two biases across the two angles per user) derived from the calibration sequence; no additional user‑specific embeddings or tokens are used.

\textbf{Objective and training.} The loss is smooth‑\(L_1\) (Huber) with outlier rejection on per‑eye angular errors between predictions and screen‑point ground truth, defined by the fixed eye origin and display geometry. Training and evaluation use identical objectives across PET and intensity baselines. Models were trained for fixed amount of 400k iterations. To isolate polarization contrast effects, baseline models are capacity‑matched (four‑channel duplicated inputs; identical backbone, head, and schedules). Models are trained per camera placement with subject‑disjoint splits.

\subsection*{Eye safety}

The near-infrared illumination complied with  IEC 62471: Photobiological safety of lamps
and lamp systems; corneal irradiance did not exceed 1.819E-02~\(\mathrm{W}\,\mathrm{m}^{-2}\) when measured 30 mm away from the eyerings with exposure time of 1 s.

\section*{Acknowledgements}

We gratefully acknowledge the contributions and support of our colleagues and collaborators throughout this project. Special thanks to Chris Wiens, Milos Micakovic, Erfan Davami, Joel Shook, Bryan Newton, Lee Garth Green, Matt Pare, Miles Thompson, Jon Battershell, Carmen Wang, Kelly Cundiff, Nicole Bell, Gavril Kochevrin, Jay-ar Kelyman, Matthew Dunbar, Alexis Terterov, Dharini Chandrashekar, Pushkar Anand, Matthias Hernandez, Raul Catena, Dylan Li, Lucas Evans, Eric Lang, Mitchell LeBlanc, Peter McEldowney, Zach Willms, Jean-Francois Parent, Andrew Anderson, ChunKiet Leong, Yueh-Tung Chen, Max Payton, Vasily Fomin, Mihika Dave, Max Ouellet, Steve Olsen, Salim Boutami, Khalid Omer, Karina Fotopoulou, Scott Campbell, Bhavana Bhoovaraghan, Mohamed El-Haddad, Rob Cavin, Scott McEldowney and Onur Akkaya.

\section*{Author contributions statement}

\textbf{Conceptualization}: M.Z., S.A., R.M., A.F., D.M.
\textbf{Methodology}: M.Z., S.A., F.A., A.B., G.P., R.M., A.F., D.M.
\textbf{Investigation}: M.Z., B.K., T.B., D.S.
\textbf{Formal analysis}: M.Z., T.B., D.S., B.K.
\textbf{Supervision and project administration}: D.S., A.F., R.M., A.B., D.M.
\textbf{Writing – original draft}: M.Z.
\textbf{Writing – review \& editing}: All authors.

\section*{Additional information}

\paragraph{Competing interests}
All authors are employees of Meta Platforms, Inc. and may hold stock and/or stock options in the company. Patents and patent applications related to polarization-enabled eye tracking and polarimetric sensing assigned to Meta Platforms, Inc., including: US Patent 12,310,660 ``Ocular hydration sensor''; US Patent App. 18/999,667 ``Polarimetric scleral position sensing for eye tracking sensor''; US Patent App. 18/951,242 ``Systems and methods for combining polarization information with time-of-flight information''; US Patent App. 18/952,477 ``Stray light reduction in eye/face tracking systems''; US Patent App. 18/367,990 ``Polarization sensitive eye authentication system''; US Patent App. 18/384,660 ``Determining and reconstructing a shape and a material property of an object''; US Patent App. 18/622,183 ``Light emitter array and beam shaping elements for eye tracking with user authentication and liveness detection''; US Patent App. 18/134,362 ``Dynamic profilometric imaging using multiscale patterns''; US Patent App. 18/413,578 ``Glucose level change detection in eyes using polarized light''; US Patent App. 18/522,874 ``Accommodation state of eye from polarized imaging''; and US Patent App. 18/071,298 ``Polarization interference pattern generation for depth sensing.'' The authors declare no other competing interests.

\bibliographystyle{assets/plainnat}
\bibliography{paper}

\end{document}